\documentclass[10pt,twocolumn,letterpaper]{article}

\usepackage{iccv}              

\definecolor{iccvblue}{rgb}{0.21,0.49,0.74}
\usepackage{graphicx}
\usepackage{multirow}
\usepackage[pagebackref,breaklinks,colorlinks,allcolors=iccvblue]{hyperref}
\usepackage{colortbl}
\usepackage{arydshln}
\usepackage[hang,flushmargin]{footmisc}
\usepackage[accsupp]{axessibility}  

\def\paperID{885} 
\def\confName{ICCV}
\def\confYear{2025}

\title{Object-centric Video Question Answering with Visual Grounding and Referring}

\author{
Haochen Wang$^{1\textbf{*}}$,\; Qirui Chen$^{2\textbf{*}}$,\; Cilin Yan$^3$,\;  Jiayin Cai$^3$,\;\\[3pt] 
Xiaolong Jiang$^3$,\; Yao Hu$^3$, Weidi Xie$^{2\dagger}$,\; Stratis Gavves$^1$
\\[3pt]
\centering{
$^{1}$University of Amsterdam \hspace{0.5cm}
$^{2}$SAI, Shanghai Jiao Tong University \hspace{0.5cm}
$^{3}$Xiaohongshu Inc. \hspace{0.5cm}
}
}

\begin{document}
\maketitle

\renewcommand{\thefootnote}{\fnsymbol{footnote}}
\footnotetext[1]{Both authors contribute equally to this work.}  
\footnotetext[2]{Corresponding author.} 

\begin{abstract}
Video Large Language Models~(VideoLLMs) have recently demonstrated remarkable progress in general video understanding. However, existing models primarily focus on high-level comprehension and are limited to text-only responses, restricting the flexibility for object-centric, multi-round interactions. 
In this paper, we make three contributions:
(i) we address these limitations by introducing a VideoLLM model, capable of performing both object referring for input and grounding for output in video reasoning tasks, {\em i.e.}, allowing users to interact with videos using both textual and visual prompts; 
(ii) we propose \textbf{STOM}~(Spatial-Temporal Overlay Module), 
a novel approach that propagates arbitrary visual prompts input at any single timestamp to the remaining frames within a video;
(iii) we present \textbf{VideoInfer}, a manually curated object-centric video instruction dataset featuring question-answering pairs that require reasoning. 
We conduct comprehensive experiments on VideoInfer and other existing benchmarks across video question answering and referring object segmentation. 
The results on 12 benchmarks of 6 tasks show that our proposed model consistently outperforms baselines in both video question answering and segmentation, underscoring its robustness in multimodal, object-centric video and image understanding. 
Project page: \href{https://qirui-chen.github.io/RGA3-release/}{https://qirui-chen.github.io/RGA3-release/}. 
\end{abstract}
    
\section{Introduction}
\label{sec:intro}

\begin{figure}[!t]
\begin{center}
\includegraphics[width=\linewidth]{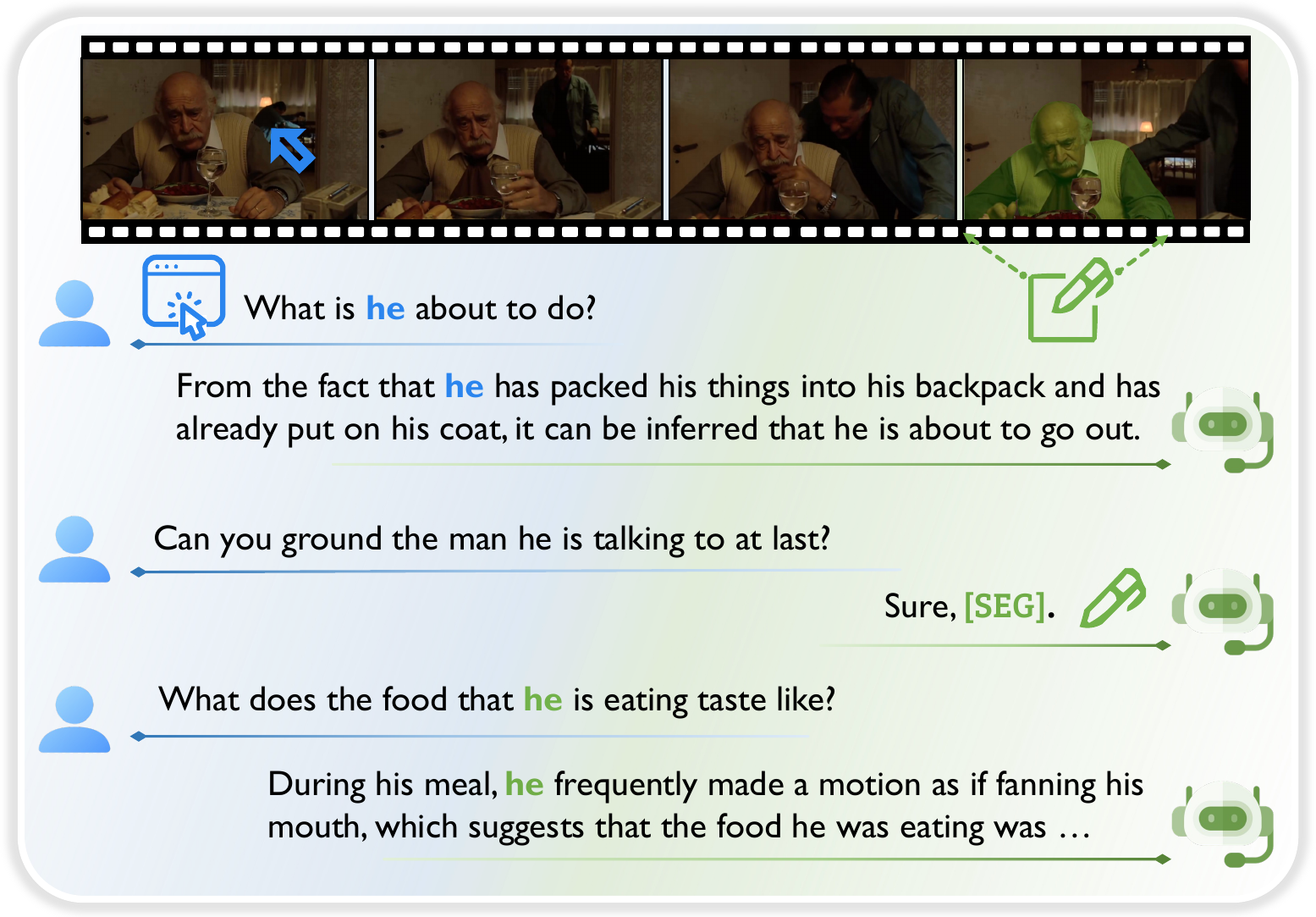}
\end{center}
\vspace{-5mm}
\caption{
\textbf{RGA3} enables both \textcolor[rgb]{0.27, 0.51, 0.91}{\textbf{referring}} and \textcolor[rgb]{0.49, 0.67, 0.33}{\textbf{grounding}} capabilities for object-centric video understanding. 
In contrast to previous VideoLLMs, which are limited to text-only responses with mask reference, RGA3 can process arbitrary \textcolor[rgb]{0.27, 0.51, 0.91}{visual prompts} and generate both textual answers and \textcolor[rgb]{0.49, 0.67, 0.33}{segmentation masks}. 
}
\label{fig:demo}
\end{figure}

Video question answering~(VideoQA) refers to the task that enables machines to understand and reason about videos. As a natural way for both humans and embodied agents~({\em e.g.}, robots or virtual assistants) to perceive the world, videos capture rich, continuous streams of data with complex spatial-temporal information, such as dynamic scenes, evolving interactions, and object behaviors. 
Despite the tremendous progress, existing Video Large Language Models~(VideoLLMs)~\cite{cheng2024videollama, li2024mvbench, yan2021videogpt, zhang2023video, lin2023video,di2024grounded,chen2024grounded} have primarily focused on holistic scene comprehension, such as answering high-level questions about videos, rather than addressing fine-grained spatial-temporal queries on objects, limiting their ability to resolve detailed object-centric reasoning in complex, dynamic video content.

This paper proposes to advance the task of multi-modal, object-centric video understanding by introducing \textbf{RGA3}~(\underline{\textbf{R}}efer and \underline{\textbf{G}}round \underline{\textbf{A}}nything \underline{\textbf{A}}nytime at \underline{\textbf{A}}ny Granularity), a unified framework designed to ground objects of interest and answer detailed questions about them in videos. 
In detail, the resulting model enables: \textbf{(i)} integrating arbitrary visual prompts—such as masks, bounding boxes, arrows, points, or scribbles—with textual inputs to perform complex reasoning tasks about specific objects in videos, and \textbf{(ii)} processing video and textual queries to generate segmentation masks for objects referred to in the query. By combining these capabilities, RGA3 facilitates object-centric reasoning, enabling more precise and interactive video understanding.

Our core innovation of \textbf{RGA3} lies in two components that enable object-centric video reasoning. The first is the Spatial-Temporal Overlay Module~(STOM), which allows the model to process arbitrary visual prompts—such as masks, bounding boxes, arrows, points, or scribbles—at any timestamp within a video. 
This enhances the model's interactivity and flexibility in handling spatial-temporal reasoning tasks. 
The second is to unify the object-centric question-answering and segmentation tasks for both images and videos based on the multimodal large language models~(MLLMs) and SAM2~\cite{ravi2024sam}, which enables the generation of segmentation masks for the referred objects in the query, 
while simultaneously reasoning over visually grounded questions. 
By integrating these two designs, \textbf{RGA3} provides a unified framework for fine-grained object grounding and reasoning in videos.

In addition to the development of model architecture, we introduce an object-centric video instruction dataset with manually crafted questions and reasoning process in the answers, namely \textbf{VideoInfer}. 
Unlike existing datasets~\cite{yuan2024videorefer}, which construct questions through automated pipelines that largely focus on object descriptions, resulting in a large proportion of perception questions, our proposed dataset emphasizes reasoning and long-term temporal understanding. VideoInfer contains around 29K question-answer (QA) pairs, and significantly surpasses existing datasets in question complexity and reasoning depth.
We conduct extensive evaluations of RGA3 on VideoInfer as well as 11 existing benchmarks, including referring visual question-answering and referring/reasoning object segmentation tasks at both image and video levels. 
The results demonstrate that RGA3 achieves state-of-the-art performance across both referring and grounding tasks. 

The rest of the paper is organized as follows: 
In Sec.~\ref{sec:related_work}, we provide an overview of recent Multi-modal Large Language Models (MLLMs) and existing approaches for referring and grounding tasks; Sec.~\ref{sec:method} details our proposed RGA3 framework and the VideoInfer dataset collection process; Sec.~\ref{sec:experiment} presents the implementation details and reports extensive experimental results and analysis on VideoInfer as well as 11 existing benchmarks.

\section{Related Work}
\label{sec:related_work}

\noindent \textbf{Multimodal Large Language Models~(MLLMs).}
Recent research in multimodal models has focused on visual instruction tuning, where pre-trained LLMs are fine-tuned for image and video understanding tasks. Models such as LLaVA~\cite{liu2024visual}, MiniGPT-4~\cite{zhu2023minigpt}, mPLUG-Owl~\cite{ye2023mplug}, Otter~\cite{li2023mimic}, and InstructBLIP~\cite{liu2024visual} have demonstrated significant progress by aligning vision-language representations through instruction tuning. Beyond text generation, recent works such as~\cite{koh2023grounding, sun2023generative, koh2024generating, liu2025LamRA} have further extended the capabilities of MLLMs to retrieval and generation tasks, showcasing the growing versatility of MLLMs.

\vspace{3pt} \noindent \textbf{Video Large Language Models~(VideoLLMs).}
Extending image-level MLLMs to the video domain presents unique challenges, since videos comprise dynamic sequences that encode temporal dependencies, requiring models to process both spatial and temporal information. 
Models such as Video-LLaVA~\cite{lin2023video}, Video-ChatGPT~\cite{yan2021videogpt}, and InternVideo~\cite{wang2022internvideo, wang2024internvideo2} have shown strong performance in tasks like video captioning, temporal reasoning, and commonsense understanding, enabling text-based interaction with video content. 
Despite these advancements, existing VideoLLMs largely focus on holistic scene understanding, generating high-level textual descriptions of videos, while failing to capture fine-grained object interactions or nuanced temporal dependencies~\cite{li2024llava, zhang2024video, li2024llava-one}. This limitation restricts their ability for object-centric reasoning, 
which is critical for detailed video analysis and practical real-world applications.

\begin{figure*}[t!]
\begin{center}
\includegraphics[width=\linewidth]{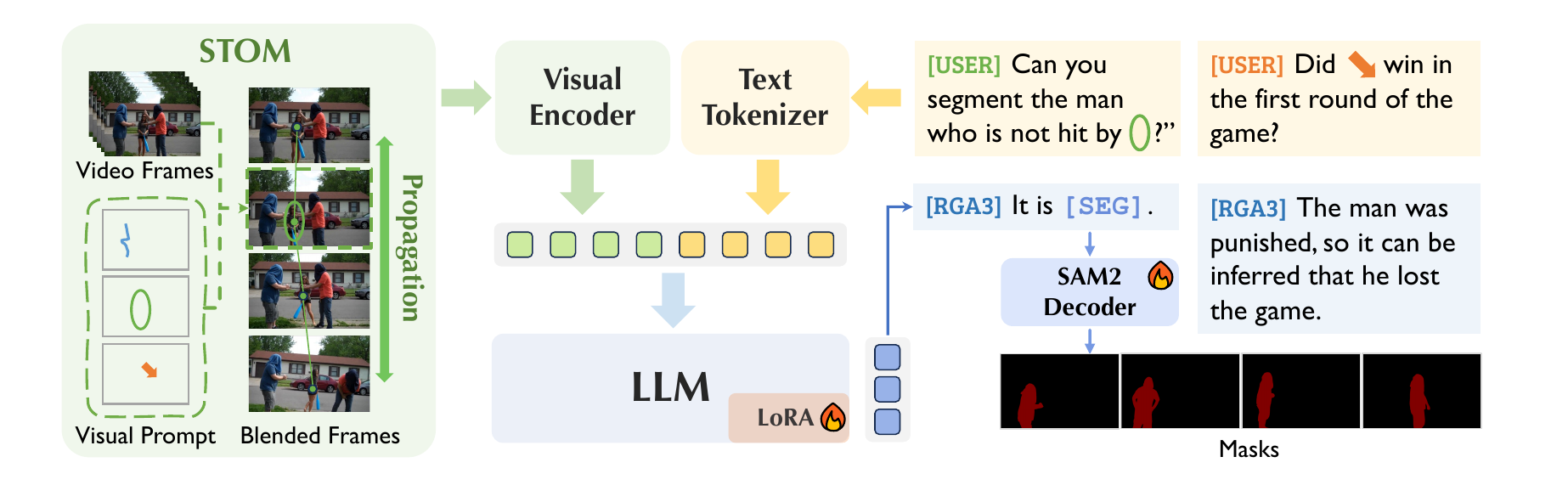}
\end{center}
\vspace{-5mm}
\caption{\textbf{The proposed RGA3 architecture overview.} (a) The \underline{S}patial-\underline{T}emporal \underline{O}verlay \underline{M}odule~(STOM) is introduced to process arbitrary visual prompts~({\em e.g.}, \textcolor{DodgerBlue}{scribble}, \textcolor{DarkGreen}{ellipse}, \textcolor{DarkOrange}{arrow}, etc.) at any timestamp and propagate to all frames, allowing for interactive object-centric reasoning and continual visual attention. (b) A visual encoder is employed to extract video representations of overlaid frames processed through STOM. (c) A Large Language Model (LLM) takes the concatenated sequence of visual and text tokens as input and generates responses. (d) To facilitate reasoning-based video object segmentation, a SAM2 decoder is incorporated for generating segmentation masks when prompted with a $\textcolor{CornflowerBlue}{\texttt{\textbf{[SEG]}}}$ token, extending RGA3’s capabilities beyond text-only responses.}
\label{fig:architecture}
\end{figure*}

\vspace{3pt} \noindent \textbf{Object-centric MLLMs.}
In the recent literature, the tasks of object-centric referring and grounding have been well-studied in the image domain, with models like GPT4ROI~\cite{zhang2023gpt4roi}, Ferret~\cite{you2023ferret}, and Osprey~\cite{yuan2024osprey} for fine-grained object understanding via region-based visual prompts. Extending these capabilities to videos is more challenging due to their temporal complexity. Recent works, such as VISA~\cite{yan2024visa}, VideoLISA~\cite{bai2025one}, and some concurrent works like VideoRefer~\cite{yuan2024videorefer}, GLUS~\cite{lin2025glus}, DAM~\cite{lian2025describe}, and Sa2VA~\cite{sa2va}, focus on either only mask-based referring question-answering or object grounding. 
In contrast, we aim to develop a unified framework for both tasks, which does not require accurate object reference~({\em i.e.,} masks) and supports arbitrary visual prompts ({\em e.g.,} boxes, points, or scribbles) at any timestamp, enabling precise, multi-round, object-centric reasoning in videos. 

\section{Method}
\label{sec:method}
In this section, we will present details on the RGA3 architecture and the dataset collection procedure for VideoInfer as follows: 
Sec.~\ref{method:problem} introduces the considered problem; 
Sec.~\ref{method:arch} details the model architecture; 
Sec.~\ref{method:dataset} describes the dataset collection process of VideoInfer, along with statistical analysis; 
Sec.~\ref{method:training} describes the instruction tuning process, in which we finetune RGA3 on VideoInfer and existing datasets through a co-training strategy.

\subsection{Problem Formulation}
\label{method:problem}
Given a video of frame sequences, our goal in this paper is to build a model $\phi_{\theta}(\cdot)$ to answer questions $\mathcal{Q}$ given an (optional) visual prompt $P_{t}$ at frame $t$:
\begin{align*}
  \mathcal{A}, \mathcal{M} = \phi_{\theta}(\mathcal{V}, \mathcal{Q}, P_{t}),
\end{align*}
where $\mathcal{V} \in \mathbb{R}^{T \times H \times W \times 3}$ is the input video containing $T$ frames, $P_t \in \mathbb{R}^{H \times W \times 4}$ is arbitrary visual prompt~(bounding boxes, masks, scribbles, arrows, or points) represented by a RGBA mark with color and alpha channels.
$\mathcal{Q}$ and $\mathcal{A}$ are questions and answers in text. 
$\mathcal{M} \in \mathbb{R}^{T \times H \times W}$ denotes the generated (optional) segmentation masks when users refer to objects in the answer.
Consequently, the model is able to process both textual input and visual prompts, jointly reasoning over spatial-temporal relationships to generate an output response, consisting of textual answers and segmentation masks to highlight the referred objects in the video, as demonstrated in Fig.~\ref{fig:demo}.

\subsection{RGA3 Architecture}
\label{method:arch}
The overall architecture of our proposed RGA3 is illustrated in Fig.~\ref{fig:architecture}. 
RGA3 employs a visual encoder to compute global video representations and a pretrained text tokenizer to get language embeddings. To facilitate region-based referring, we propose STOM, a \underline{S}patial-\underline{T}emporal \underline{O}verlay \underline{M}odule that propagates arbitrary visual prompts input at any timestamp to the remaining frames for long-term video object referring.
Additionally, we incorporate the SAM2~\cite{ravi2024sam} decoder to generate segmentation masks, extending RGA3’s capabilities beyond text-only responses.

\vspace{3pt} \noindent \textbf{Visual Prompting.}
A visual prompt $P_t$, {\em e.g.}, masks, boxes, or scribbles at timestamp $t$ is represented as an RGBA image $P_t \in \mathbb{R}^{H \times W \times 4}$, where the mark is colored, and the background is opaque. 
STOM propagates visual prompt $P_t$ at frame $t$ to all frames $\{P_1, P_2, \ldots, P_T\}$ to get comprehensive object referring. Specifically, we adopt an off-the-shelf point tracker~\cite{karaev2024cotracker3} to track points around the centroid of the visual prompt bi-directionally.
The pixels to be tracked are defined as $\{\mathbf{p} \in \mathbb{R}^2 \mid |\mathbf{p} - \mathbf{c}|_2 \leq r \}$, where $\mathbf{c} \in \mathbb{R}^2$ is the geometric center of $P_t$. 
A new circle (or the original shape) covering visible propagated pixels is drawn for each propagated frame, preserving the visual prompt’s color and opacity. 
We use \textbf{alpha blending}~($\Phi(\cdot)$) to overlay the visual prompts $\{P_1, P_2, \ldots, P_T\}$ onto the original video frames: $\mathcal{V}'_i =\Phi(\mathcal{V}_i, P_i),\; \forall 1\leq i \leq T$. $\mathcal{V}'_i$ is then fed to multimodal large language model~(MLLM) in the following stage.

\begin{figure*}[!th]
\begin{center}
\includegraphics[width=\linewidth]{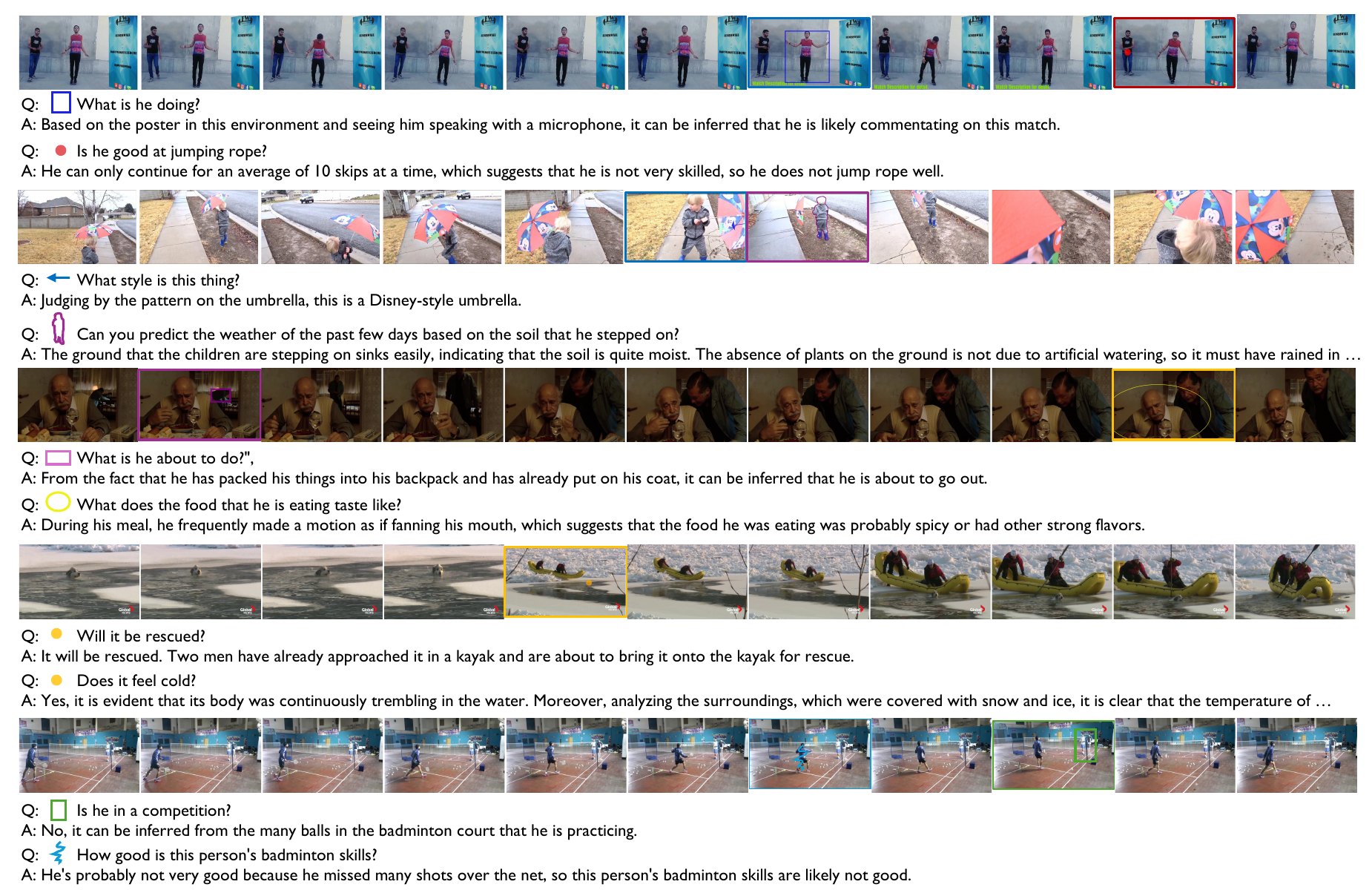}
\end{center}
\vspace{-5mm}
\caption{\textbf{Visualization of samples in VideoInfer dataset.} Some long answers have been truncated to accommodate width constraints.}
\label{fig:vis_video_infer_samples}
\end{figure*}

Unlike mask pooling or region-based feature extraction, alpha blending avoids pooling out unstable features, especially for prompts like arrows or circles that may only partially overlap with target objects. Additionally, by blending prompts onto frames, the model can directly learn geometric relationships between prompts and objects, enhancing robustness in object referring.

\vspace{2pt} \noindent \textbf{Video-Language Input.}
Blended video frames $\mathcal{V}'$ are processed using the pre-trained visual encoder from Qwen2.5-VL~\cite{Qwen2.5-VL}, which extracts visual features from $T$ frames and projects them into embeddings $v_{\text{visual}} \in \mathbb{R}^{\frac{T}{2} \times \frac{H}{14} \times \frac{W}{14} \times D}$. 
Text input is tokenized and encoded into embeddings $v_{\text{text}} \in \mathbb{R}^{L_t \times D}$. 
The visual and text embeddings are then concatenated and fed into the LLM, enabling joint reasoning over visual and textual modalities for object-centric video understanding. 
For segmentation outputs, frames are additionally resized and encoded into dense feature maps $v_{\text{dense}} \in \mathbb{R}^{T \times H^{\prime} \times W^{\prime} \times C}$ using the Hiera-L~\cite{ryali2023hiera} encoder from SAM2~\cite{ravi2024sam}~(omitted in Fig.~\ref{fig:architecture} for simplicity), triggered when a $\texttt{<SEG>}$ token is detected in the response.

\vspace{2pt} \noindent \textbf{Decoder for Language and Grounding.}
LLM generates a sequence of text tokens, in which we add a special token $\texttt{<SEG>}$ for segmentation outputs.
Specifically, we locate the $\texttt{<SEG>}$ tokens in text output, and project the corresponding last layer's hidden states into the embedding space of the SAM2 prompt encoder, which is denoted as $\mathbf{h}\in\mathbb{R}^C$.
The embedding $\mathbf{h}$ are then passed to the SAM2~\cite{ravi2024sam} decoder $\Psi_{\text{dec}}(\cdot)$ with dense visual feature maps $v_{\text{dense}}$, to produce the final segmentation masks $\mathcal{M} \in \mathbb{R}^{T \times H \times W}$:
\begin{align*}
    \mathcal{M} = \Psi_{\text{dec}}(\mathbf{h}, v_{\text{dense}})
\end{align*}
This integration allows the model to seamlessly combine language understanding and visual grounding, enabling joint reasoning and grounding outputs.

\subsection{VideoInfer Dataset}
\label{method:dataset}

This section presents the details of \textbf{VideoInfer}, 
a manually curated, object-centric video question-answering dataset, designed to challenge models with questions requiring semantic understanding, temporal reasoning, and instance discrimination over video content. 
Compared to existing object-level video question-answering datasets, which are often generated through automated pipelines, VideoInfer serves as a more rigorous benchmark for evaluating the reasoning capabilities of advanced VideoLLMs.

\begin{figure*}[t!]
\begin{center}
\includegraphics[width=\linewidth]{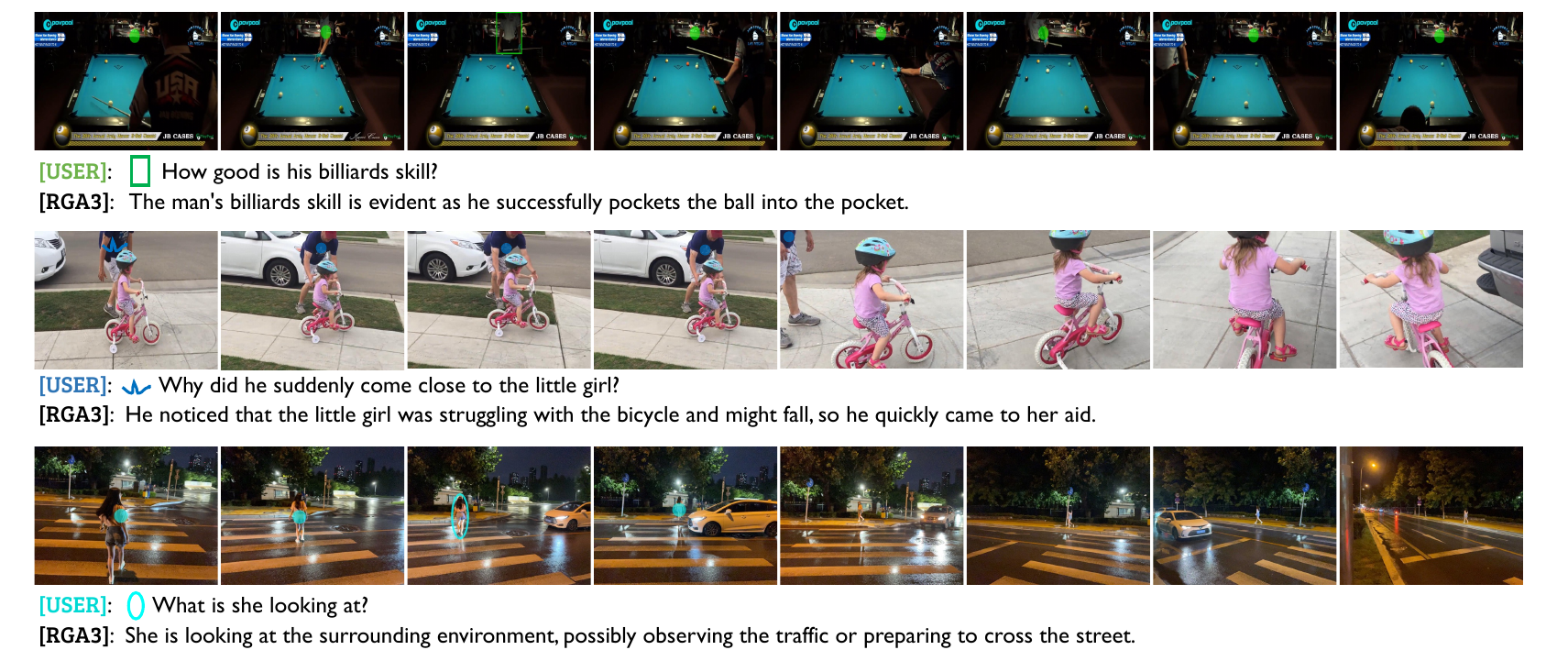}
\end{center}
\vspace{-5mm}
\caption{\textbf{Visualization of RGA3 results on VideoInfer dataset}~(zoom in for a better view).}
\label{fig:vis_videoinfer}
\end{figure*}

\vspace{3pt} \noindent \textbf{Dataset Collection.}
To construct VideoInfer, we curated 1,650 videos from diverse, high-quality object-centric video datasets, including TAO~\cite{athar2023burst}, LVOS~\cite{hong2024lvos}, LVVIS~\cite{wang2024ov}, VIPSeg~\cite{miao2022large}, 
UVO~\cite{wang2021unidentified}, MOSE~\cite{ding2023mose}, and OVIS~\cite{qi2022occluded}, which cover diverse scenarios such as indoor/outdoor activities, animal behaviors, and human interactions.
These datasets provide dense mask annotations for objects that are used to automatically generate additional types of visual prompts, such as boxes, scribbles, and arrows following~\cite{cai2024vip}.

We prioritized long-form videos from TAO and LVOS for their rich, temporally dynamic content, making them ideal for crafting implicit reasoning questions. 
The original datasets provide masks or bounding boxes as annotations.
We randomly generate 8 types of visual prompts (mask,
mask contour, rectangle, ellipse, triangle, scribble, arrow, and point) based on mask annotations, and only prompt 1 frame for each video, to mimic the operations of the users.
Each object with the visual prompt is manually annotated with question-answering (QA) pairs, following rules: \textbf{(i) indirectness:} the answer cannot be directly observed from the video but requires understanding and reasoning about the object of interest from users; or \textbf{(ii) dynamism:} questions require temporal reasoning, making it impossible to answer from a single frame.
We apply cross-validation among two annotation groups, and low-scoring question-answer pairs are subsequently discarded.

\vspace{3pt} \noindent \textbf{Statistical Analysis.}
VideoInfer contains 1,620 videos and 28,811 QA pairs, split into two subsets: an instruction tuning set (950 videos with 2,555 objects and 20,320 QA pairs) and a challenging test set (670 videos with 1,572 objects and 8,491 QA pairs). 
This division provides a rigorous benchmark for evaluating future VideoLLMs.
Visualizations of VideoInfer samples are provided in Fig.\ref{fig:vis_video_infer_samples}, 
most of the questions require inference and reasoning instead of directly observing the answer from the video.
For example, in the first case, the question ``Is he good at jumping rope?'' requires the model to go through the entire video and analyze the actions of the man. 
Comparing with VideoRefer-700K~\cite{yuan2024videorefer}, 
that contains automatically generated multiple-choice QA, 
our VideoInfer dataset contains open-ended questions that require deep reasoning.

\subsection{Instruction Tuning}
\label{method:training}

We finetune \textbf{RGA3} with a unified image-video training pipeline, treating images as single-frame videos to ensure smooth adaptation across both modalities. 
A complete list of training datasets is provided in Sec.~\ref{sec:exp_setting}. 
Since most existing Referring QA datasets provide only masks or bounding box annotations, we augment them by randomly generating different types of visual prompts during training. 
This data augmentation improves RGA3's ability to handle diverse and arbitrary visual prompts effectively.

To enable both textual responses and segmentation outputs, we adopt a multi-task learning objective. 
Specifically, the text generation loss is defined as an auto-regressive cross-entropy loss~($\mathcal{L}_{\text{txt}}$), 
while the segmentation loss~($\mathcal{L}_{\text{mask}}$) combines per-pixel binary cross-entropy (BCE) loss and DICE loss~\cite{milletari2016v}. Given the ground-truth targets ($\hat{\mathbf{y}}_{\text{txt}}$, $\mathbf{\hat{m}}_{\text{tgt}}$) and the predictions $(\mathbf{y}_{\text{txt}}$, $\mathbf{m}_{\text{tgt}}$), the training objectives can be formulated as:
\begin{align*}
   &\mathcal{L}_{\text{txt}} = \text{CE}(\hat{\mathbf{y}}_{\text{txt}}, \mathbf{y}_{\text{txt}}) \\
   &\mathcal{L}_{\text{mask}} = \lambda_{\text{bce}} \cdot \text{BCE}(\mathbf{\hat{m}}_{\text{tgt}}, \mathbf{m}_{\text{tgt}}) + \lambda_{\text{dice}} \cdot \text{DICE}(\mathbf{\hat{m}}_{\text{tgt}}, \mathbf{m}_{\text{tgt}})
\end{align*}
The overall objective $\mathcal{L}$ is the weighted sum of the text generation and segmentation losses: $\mathcal{L}=\lambda_{\text{txt}} \cdot \mathcal{L}_{\text{txt}} + \mathcal{L}_{\text{mask}}$.
\section{Experiment}
\label{sec:experiment}

In this section, we first provide a detailed overview of the training datasets, implementation details, and evaluation metrics in Sec.~\ref{sec:exp_setting}. 
Then, we present extensive comparisons between RGA3 and state-of-the-art methods across a variety of visual QA and referring object segmentation benchmarks in Sec.~\ref{sec:comparison}. 
Finally, we conduct ablation studies to analyze the impact of STOM and SAM2 modules, indicating their contributions to overall performance in Sec.~\ref{sec:ablation}.

\subsection{Experimental Setting}
\label{sec:exp_setting}

\vspace{3pt} \noindent \textbf{Training Datasets.}
The training datasets of RGA3 consist of two main parts: 
image/video QA datasets and image/video segmentation datasets. 
\textbf{(i)} We utilize various image-based QA datasets to improve the general question-answering capabilities, along with video-based QA datasets to foster temporal reasoning and inference;
\textbf{(ii)} we include image/video segmentation datasets, which contain general semantic image segmentation datasets, and referring/reasoning object segmentation datasets to enable the grounding ability for RGA3.
To enhance interactive visual reasoning, we mix both QA datasets and segmentation datasets during training. 
The detailed composition of the training datasets is listed in the Supplementary Material.

Specifically, for the referring QA datasets, all the object references are initially annotated as masks or bounding boxes.
We then randomly generate versatile visual prompts (mask, mask contour, rectangle, ellipse, triangle, scribble, arrow, and point) based on mask and bounding box annotations for both a single image and video frames, enabling more natural and flexible user interactions in instruction tuning. 
This visual prompt generation process is conducted on all referring QA datasets, including ViP-LLaVA-Instruct~\cite{cai2024vip}, Osprey-724K~\cite{yuan2024osprey}, and proposed VideoInfer.

\begin{table}[t]
\setlength{\tabcolsep}{1.5pt}
\centering
{
\scalebox{0.85}{
\begin{tabular}{lccccc}
\toprule
\multirow{2}{*}{\textbf{Method}} 
 & \multirow{2}{*}{BLEU-4} & \multirow{2}{*}{CIDEr} & \multirow{2}{*}{ROUGE-L} & \textbf{GPT-4o} \\
 & & & & (\textbf{Acc. / Score}) \\
\hline
\rowcolor{gray!10}
\addlinespace[0.1em]
\multicolumn{5}{c}{\textit{Generalist Models}} \\
\addlinespace[0.2em]
VideoLLaMA3-7B~\cite{zhang2025videollama} & 4.1 & 48.0 & 21.2 & 39.1 / 2.28\\
Qwen2.5-VL-3B~\cite{Qwen2.5-VL}  & 6.5 & 53.8 & 19.0 & 28.4 / 2.14 \\
Qwen2.5-VL-7B~\cite{Qwen2.5-VL}  & 6.4 & 54.6 & 23.8 & 41.7 / 2.42 \\
GPT-4o~(detail:low)~\cite{hurst2024gpt} & 7.6 & 62.7 & 22.9 & 44.1 / 2.48 \\
GPT-4o~(detail:high)~\cite{hurst2024gpt}  & 7.6 & 64.4 & 23.0 & 45.5 / 2.50 \\
\hline
\rowcolor{gray!10}
\addlinespace[0.1em]
\multicolumn{5}{c}{\textit{Specialist Models}} \\
\addlinespace[0.2em]
Osprey-7B~\cite{yuan2024osprey} & 4.8 & 44.6 & 20.2 & 27.5 / 1.67 \\
VideoRefer-7B~\cite{yuan2024videorefer} & 7.4 & 77.8 & 25.6 & 40.4 / 2.29 \\
\arrayrulecolor{gray}\hdashline
\addlinespace[0.3em]
 \rowcolor{LightSkyBlue!20}
\textbf{RGA3-7B} & 12.2 & 104.3 & 29.9 & \textbf{46.7 / 2.60} \\
\bottomrule
\end{tabular}
}
}
\vspace{-2mm}
\caption{Evaluation of open-ended referring VideoQA task on our \textbf{VideoInfer} benchmark, given 8 different types of visual prompts.
The `detail' parameter adjusts the resolution of GPT-4o input.
}
\label{tab:video_infer}
\end{table}

\vspace{3pt} \noindent \textbf{Evaluation Metrics.}
We evaluate our model from two primary perspectives: Referring QA and Referring/Reasoning object segmentation at both the video-level and image-level.
\textbf{(i)} For video Referring QA, we report accuracy on multiple-choice questions on VideoRefer-Bench\textsuperscript{Q}~\cite{yuan2024videorefer}. 
On our VideoInfer benchmark, we adopt BLEU-4~\cite{papineni2002bleu}, CIDEr~\cite{vedantam2015cider}, ROUGE-L~\cite{lin2004rouge} and GPT-4o evaluated accuracy\,/\,score~\cite{maaz2023video} as metrics for open-ended questions, following previous works~\cite{di2024grounded,maaz2023video}.
For image Referring QA, we follow the official GPT-4-turbo-based evaluation on ViP-Bench~\cite{cai2024vip}, averaging results over 5 runs. 
\textbf{(ii)} For Referring/Reasoning video object segmentation, we adopt region similarity ($\mathcal{J}$), contour accuracy ($\mathcal{F}$), and their average ($\mathcal{J\&F}$), consistent with prior methods. 
The robustness score ($\mathcal{R}$)~\cite{li2022towards} is adopted to evaluate hallucinations on ReVOS~\cite{yan2024visa}.

\begin{table}[t]
\setlength{\tabcolsep}{3.0pt}
\centering
\scalebox{0.88}{
\begin{tabular}{lccccccc}
\toprule
\textbf{Method} & \textbf{Input} & Basic & Seq. & Rel. & Rea. & Pred. & \textbf{Avg.} \\ 
\midrule
\rowcolor{gray!10}
\multicolumn{8}{c}{\textit{Generalist Models}} \\
\addlinespace[0.2em]
Qwen2-VL-7B~\cite{wang2024qwen2} & SoM & 62.0 & 69.6 & 54.9 & 87.3 & 74.6 & 66.0 \\
InternVL2-26B~\cite{chen2024far} & SoM & 58.5 & 63.5 & 53.4 & 88.0 & 78.9 & 65.0 \\
GPT-4o-mini~\cite{hurst2024gpt} & SoM & 57.6 & 67.1 & 56.5 & 85.9 & 75.4 & 65.8 \\
GPT-4o~\cite{hurst2024gpt} & SoM & 62.3 & 74.5 & 66.0 & 88.0 & 73.7 & 71.3 \\
\hline
\rowcolor{gray!10}
\addlinespace[0.1em]
\multicolumn{8}{c}{\textit{Specialist Models}} \\
\addlinespace[0.2em]
Osprey-7B~\cite{yuan2024osprey} & RF & 45.9 & 47.1 & 30.0 & 48.6 & 23.7 & 39.9 \\
Ferret-7B~\cite{you2023ferret} & RF & 35.2 & 44.7 & 41.9 & 70.4 & 74.6 & 48.8 \\
VideoRefer-7B~\cite{yuan2024videorefer} & RF & 75.4 & 68.6 & 59.3 & 89.4 & 78.1 & 71.9 \\
\arrayrulecolor{gray}\hdashline
\addlinespace[0.3em]
 \rowcolor{LightSkyBlue!20}
\textbf{RGA3-7B~(Ours)} & \textbf{VP} & 77.4 & 71.9 & 61.1 & 88.8 & 81.6 & \textbf{74.0} \\ 
\bottomrule
\end{tabular}
}

\vspace{-2mm}
\caption{Evaluation of multiple-choice referring VideoQA task on \textbf{VideoRefer-Bench$^\text{Q}$}~\cite{yuan2024videorefer}, given different object referring approaches.
`SoM', `RF', and `VP' represent Set-of-Mask~\cite{yang2023set}, region feature, and visual prompt, respectively.}
\label{tab:video_refer}
\end{table}

\vspace{3pt} \noindent \textbf{Implementation Details.}
Our proposed RGA3 model is implemented with Qwen2.5-VL-7B~\cite{Qwen2.5-VL} as the visual-language backbone and SAM2 (Hiera-L)~\cite{ravi2024sam, ryali2023hiera} as the grounding module by default. 
The training is conducted on 16 NVIDIA H800 (80GB) GPUs distributed across 2 nodes, with a batch size of 2 per device and gradient accumulation steps set to 8, leading to a total of 256 effective samples per update. 
We utilize the DeepSpeed engine to manage distributed training and optimization efficiently.
The optimization process employs the AdamW optimizer~\cite{loshchilov2017decoupled}, combined with a WarmupCosineLR learning rate scheduler. 
The maximum learning rate is set to $4\times10^{-5}$, with a gradual warm-up phase to stabilize initial training dynamics. 
We assign the loss weights for the next token prediction loss, binary cross-entropy (BCE) loss, and DICE loss as 1.0, 2.0, and 0.5, respectively, ensuring a balanced training objective for both text generation and segmentation mask prediction.
Throughout the training process, the visual encoders remain frozen to preserve pre-trained visual knowledge, while the LLM is fine-tuned using LoRA~\cite{hu2021lora}, enabling efficient and memory-friendly adaptation of the model parameters.

\begin{table}[t]
\vspace{10pt}
\centering
\resizebox{0.48\textwidth}{!}{
\begin{tabular}{l|c|cc}
\toprule
\multirow{1}{*}{\textbf{Method}} & \multirow{1}{*}{\textbf{Format}} &  \textbf{Synthesized} & \textbf{Manual}  \\
 \midrule
GPT-4V-turbo~(detail: high)~\cite{achiam2023gpt} & VP & 60.7 & 59.9 \\
GPT-4V-turbo~(detail: low)~\cite{achiam2023gpt} & VP & 52.8 & 51.4 \\
\midrule
Shikra-7B~\cite{chen2023shikra} & Coor & 33.7 & – \\
GPT4ROI-7B~\cite{zhang2023gpt4roi} & ROI & 35.1 & – \\
Kosmos-2~\cite{peng2023kosmos} & Dis & 26.9 & – \\
ViP-LLaVA-Base-13B~\cite{cai2024vip} & VP & 48.2 & 47.0 \\
ViP-LLaVA-13B~\cite{cai2024vip} & VP & 48.3 & 48.2 \\
\arrayrulecolor{gray}\hdashline
\addlinespace[0.3em]
 \rowcolor{LightSkyBlue!15}
\textbf{RGA3-7B~(Ours)} & VP & \textbf{48.7} & \textbf{49.4} \\
\bottomrule
\end{tabular}
}
\vspace{-2mm}
\caption{Evaluation of image-level referring QA task on \textbf{ViP-Bench}~\cite{cai2024vip}, given eight different types of visual prompts. 
}
\label{tab:vip}
\end{table}

\begin{figure*}[t!]
\begin{center}
\includegraphics[width=\linewidth]{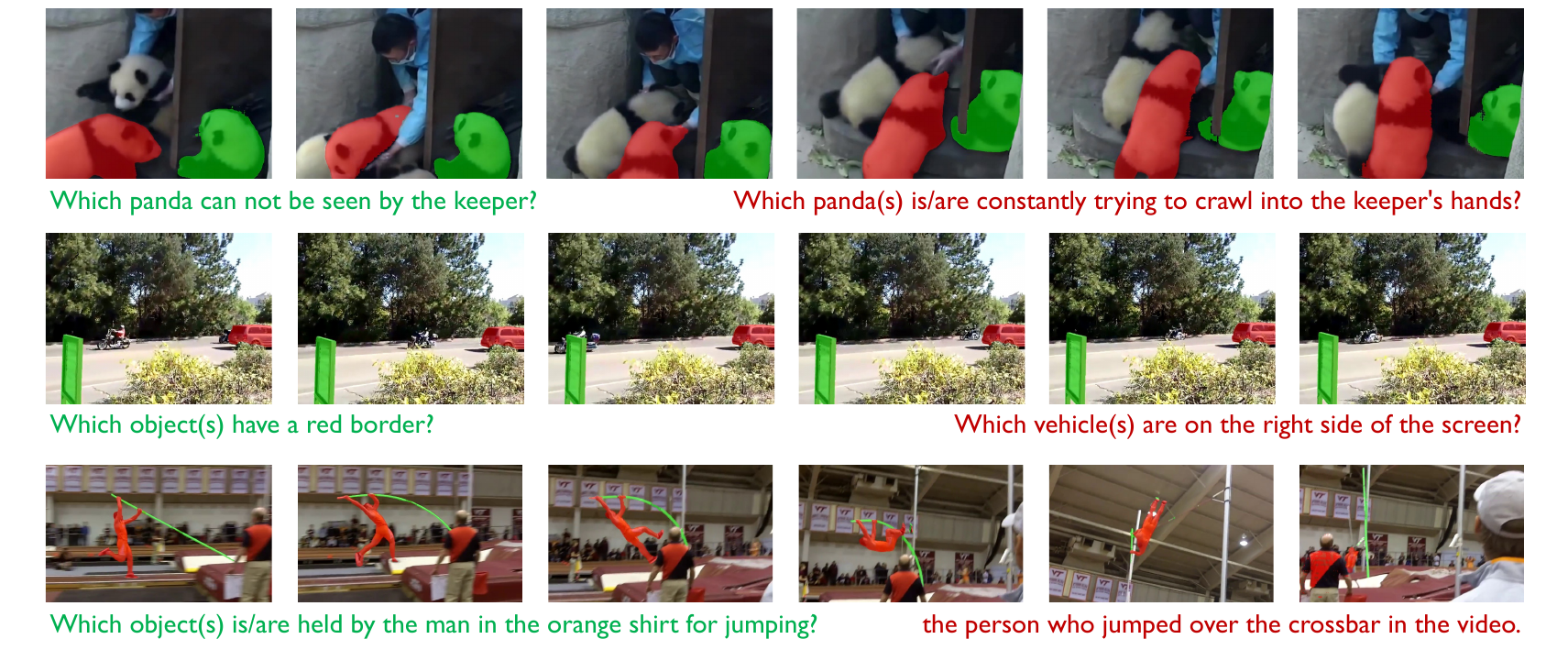}
\end{center}
\vspace{-5mm}
\caption{Visualization of the RGA3 grounding results on \textbf{ReVOS} dataset.}
\label{fig:vis_revos}
\end{figure*}

\begin{table*}[ht]
\setlength{\tabcolsep}{7.0pt}
\centering
{
\footnotesize
\begin{tabular}{l|ccc|cccccccccc}
\toprule
\multirow{3}{*}{\textbf{Method}} & \multicolumn{3}{c|}{\multirow{2}{*}{\textbf{ReasonVOS}~\cite{bai2025one}}} & \multicolumn{10}{c}{\textbf{ReVOS}~\cite{yan2024visa}} \\ 
\cmidrule(lr){5-14}
 & \multicolumn{3}{l|}{} & \multicolumn{3}{c|}{overall} & \multicolumn{3}{c|}{referring} & \multicolumn{3}{c|}{reasoning} & \multirow{2}{*}{$\mathcal{R}$} \\ 
 \cmidrule(lr){2-4} \cmidrule(lr){5-7} \cmidrule(lr){8-10} \cmidrule(lr){11-13}
 & $\mathcal{J\&F}$ & $\mathcal{J}$ & $\mathcal{F}$ & $\mathcal{J\&F}$ & $\mathcal{J}$ & \multicolumn{1}{l|}{$\mathcal{F}$} & $\mathcal{J\&F}$ & $\mathcal{J}$ & \multicolumn{1}{l|}{$\mathcal{F}$} & $\mathcal{J\&F}$ & $\mathcal{J}$ & \multicolumn{1}{l|}{$\mathcal{F}$} &  \\ 
 \midrule
LISA-7B~\cite{lai2024lisa} & 31.1 & 29.1 & 33.1 & 40.9 & 39.1 & \multicolumn{1}{l|}{42.7} & 45.7 & 44.3 & \multicolumn{1}{l|}{47.1} & 36.1 & 33.8 & \multicolumn{1}{l|}{38.4} & 9.3 \\
VISA-7B~\cite{yan2024visa} & - & - & - & 46.9 & 44.9 & \multicolumn{1}{l|}{49.0} & 50.9 & 49.2 & \multicolumn{1}{l|}{52.6} & 43.0 & 40.6 & \multicolumn{1}{l|}{45.4} & 15.5 \\ 
VideoLISA-3.8B~\cite{bai2025one} & 45.1 & 43.1 & 47.1 & 48.4 & 46.1 & \multicolumn{1}{l|}{50.7} & 53.0 & 51.0 & \multicolumn{1}{l|}{55.1} & 43.8 & 41.2 & \multicolumn{1}{l|}{46.4} & 20.1 \\
 \midrule
 \rowcolor{LightSkyBlue!15}
RGA3-3B~(Ours) & 51.7  &  49.1 & 54.3  & 56.1 & 54.1 & \multicolumn{1}{l|}{58.0} & 59.3 & 57.6 & \multicolumn{1}{l|}{61.0} & 52.8 & 50.6 & \multicolumn{1}{l|}{55.0} &  27.7 \\
\rowcolor{LightSkyBlue!15}
RGA3-7B~(Ours) & \textbf{53.6} & \textbf{51.3} & \textbf{56.0} & \textbf{58.0} & \textbf{55.9} & \multicolumn{1}{l|}{\textbf{60.0}} & \textbf{60.5} & \textbf{58.7} & \multicolumn{1}{l|}{\textbf{62.3}} & \textbf{55.4} & \textbf{53.1} & \multicolumn{1}{l|}{\textbf{57.7}} & \textbf{28.6} \\ 
\bottomrule
\end{tabular}
}
\vspace{-3mm}
\caption{Comparison on video-level reasoning object segmentation datasets.}
\label{tab:video_reason_seg}
\end{table*}

\subsection{Comparison}
\label{sec:comparison}
\vspace{3pt} \noindent \textbf{Referring Question Answering.} We evaluate RGA3 on both image and video-level referring QA datasets, including VideoInfer~(Ours), VideoRefer-Bench\textsuperscript{Q}~\cite{yuan2024videorefer}, and VIP-Bench~\cite{cai2024vip}. 
The performance comparison on VideoInfer is presented in Tab.~\ref{tab:video_infer}.
As shown, RGA3 outperforms the concurrent work VideoRefer~\cite{yuan2024videorefer} by 6.3\,/\,0.31 in terms of GPT-4o evaluated accuracy\,/score, demonstrating its strong ability in video reasoning and object-centric understanding under various visual prompts.
As VideoInfer contains different types of visual prompts instead of pure masks as object referring inputs, this comparison indicates the strong ability of RGA3 to process arbitrary visual prompts, and thus being more interactive friendly.

Results in Tab.~\ref{tab:video_refer} show that RGA3 achieves state-of-the-art performance across both image-level and video-level referring QA tasks. 
Specifically, RGA3-7B surpasses the best region feature-based method VideoRefer-7B~\cite{yuan2024videorefer} by 2.1\% and exceeds the SoM-based GPT-4o method by 2.7\% on the VideoRefer-Bench\textsuperscript{Q} dataset. 
As shown in Tab.~\ref{tab:vip}, without being specifically designed for image-level tasks, RGA3 achieves comparable performance with ViP-LLaVA-13B models~\cite{cai2024vip} on ViP-Bench~\cite{cai2024vip} for image-level referring QA with diverse visual prompt types.
The visualization of RGA3 on the VideoInfer benchmark is shown in Fig.~\ref{fig:vis_videoinfer}.

\begin{table*}[th]
\setlength{\tabcolsep}{8.0pt}
\centering
{
\footnotesize
\begin{tabular}{l|ccc|ccc|ccc|ccc}
\toprule
\multirow{2}{*}{\textbf{Method}} & \multicolumn{3}{c|}{\textbf{MeViS} \textit{val}$^u$ ~\cite{ding2023mevis}} & \multicolumn{3}{c|}{\textbf{MeViS} \textit{val}~\cite{ding2023mevis} } & \multicolumn{3}{c|}{\textbf{Ref-YouTube-VOS}~\cite{seo2020urvos}} & \multicolumn{3}{c}{\textbf{Ref-DAVIS-17}~\cite{khoreva2019video}} \\ \cmidrule(lr){2-4} \cmidrule(lr){5-7} \cmidrule(lr){8-10} \cmidrule(lr){11-13} 
 & $\mathcal{J\&F}$ & $\mathcal{J}$ & $\mathcal{F}$ & $\mathcal{J\&F}$ & $\mathcal{J}$ & $\mathcal{F}$ & $\mathcal{J\&F}$ & $\mathcal{J}$ & $\mathcal{F}$ & $\mathcal{J\&F}$ & $\mathcal{J}$ & $\mathcal{F}$ \\ 
 \midrule
LISA-7B~\cite{lai2024lisa} & 44.8 & 41.1 & 48.6 & 37.2 & 35.1 & 39.4 & 50.2 & 49.7 & 50.6 & 58.4 & 54.9 & 61.9 \\
VISA-7B~\cite{yan2024visa} & - & - & - & 43.5 & 40.7 & 46.3 & 61.5 & 59.8 & 63.2 & 69.4 & 66.3 & 72.5 \\
VideoLISA-3.8B~\cite{bai2025one} & 51.7 & 48.4 & 57.9 & 44.4 & 41.3 & 47.6 & 61.7 & 60.2 & 63.3 & 67.7 & 63.8 & 71.5 \\ 
 \midrule
 \rowcolor{LightSkyBlue!15}
RGA3-3B~(Ours) & 57.7  & 54.7  & 60.7  &  48.8  &  46.2  & 51.5 & 67.4  & 65.8   & 69.1   &  72.1  &  67.6  &  76.6 \\
 \rowcolor{LightSkyBlue!15}
RGA3-7B~(Ours) & \textbf{59.7} & \textbf{56.7} & \textbf{62.6} & \textbf{50.1} & \textbf{47.4} & \textbf{52.8} & \textbf{68.5}  & \textbf{66.8} & \textbf{70.1} & \textbf{72.8} & \textbf{68.3} & \textbf{77.3} \\ 
\bottomrule
\end{tabular}
}
\vspace{-1mm}
\caption{Comparison on video-level referring object segmentation datasets. All comparisons on segmentation tasks are conducted under the same end-to-end manner without extra post-optimization methods such as XMem~\cite{cheng2022xmem} or XMem++~\cite{bekuzarov2023xmem++}.}
\label{tab:video_ref_seg}
\end{table*}

\vspace{3pt} \noindent \textbf{Referring \& Reasoning Segmentation.} 
As shown in Tab.~\ref{tab:video_reason_seg} and Tab.~\ref{tab:video_ref_seg}, RGA3 significantly outperforms the existing state-of-the-art video object segmentation models across multiple benchmarks.
Specifically, RGA3-7B surpasses state-of-the-art MLLM-based methods VISA~\cite{yan2024visa} and VideoLISA~\cite{bai2025one} by around 10\% on both the ReasonVOS~\cite{bai2025one} and ReVOS~\cite{yan2024visa} datasets, demonstrating its superior reasoning-based video object segmentation ability. 
On general referring video object segmentation datasets, including MeVIS~\cite{ding2023mevis}, Ref-Youtube-VOS~\cite{seo2020urvos}, and Ref-DAVIS-17~\cite{khoreva2019video}, RGA3 consistently outperforms existing VideoLLMs~(VideoLISA~\cite{bai2025one}, VISA~\cite{yan2024visa}) and image-level MLLMs such as LISA~\cite{lai2024lisa}, highlighting its robust grounding capability across diverse video segmentation datasets.
The visualization results of RGA3 on the video reasoning segmentation benchmark are presented in Fig.~\ref{fig:vis_revos}. 
Moreover, RGA3 outperforms state-of-the-art MLLM-based methods on image-level segmentation benchmarks including RefCOCO(+/g)~\cite{kazemzadeh2014referitgame,mao2016generation} and ReasonSeg~\cite{lai2024lisa}, demonstrating the general grounding ability of RGA3 on image-level grounding tasks. 
The detailed results on image segmentation benchmarks are shown in the Supplementary Material.

In summary, our proposed RGA3 achieves state-of-the-art performance in both video referring QA and video referring/reasoning object segmentation tasks, enabling multi-modal processing in both model inputs and outputs. 
This advancement enhances user interaction by supporting multi-round and multi-modal conversational object-centric reasoning on videos.

\subsection{Ablation Studies}
\label{sec:ablation}
\vspace{3pt} \noindent \textbf{Ablation of the proposed STOM module.} 
In Tab.~\ref{ablation:prop}, we investigate the effectiveness of STOM. 
The region feature-based approach computes average pixel features through mask pooling, which serves as a common strategy for region-level reasoning.
However, an arrow or scribble pointing to an object may only partially overlap with the object of interest, leading to a noisy and incomplete object feature.
To overcome this limitation, we adopt an overlay-based approach that directly integrates visual prompts with original frames, allowing the model to learn end-to-end geometric relationships between different visual prompt types and their corresponding objects. 
As shown, RGA3 w/o STOM achieves 43.6\,/\,2.46 on the VideoInfer dataset.
By incorporating tracking and overlaying, RGA3 w/ STOM achieves 46.7\,/\,2.60, demonstrating the effectiveness of our overlay-based approach for spatial-temporal prompt reasoning.
The oracle propagation in the last row refers to the use of ground truth masks to generate consistent visual prompts for all frames as the object reference. 

\vspace{2pt} \noindent \textbf{Ablation of the segmentation module.} 
As shown in Tab.~\ref{supp:sam}, RGA3-7B with the SAM decoder~\cite{kirillov2023segment} already outperforms VISA-7B~\cite{yan2024visa}~(which uses the same SAM decoder) by 8.5 and 15.7 on the ReVOS and ReasonSeg benchmarks, respectively. 
Incorporating SAM2 provides a 2.6 improvement on ReVOS~\cite{yan2024visa} and a marginal gain on ReasonSeg~\cite{lai2024lisa} compared to the SAM decoder, 
suggesting that RGA3’s segmentation performance improvements stem primarily from its comprehensive video content understanding and memory attention mechanism. 

We also provide the latency and memory usage of the STOM and SAM2 module on 1K samples of VideoRefer-Bench$^{Q}$ / ReVOS with one H800 (80G) in tab.~\ref{tab:speed}. 
We find that the propagation process of SAM2 and the tracking process of CoTracker3 cost most time, except for the forward process of LLM.

\begin{table}[t]
\vspace{5pt}
\centering
\footnotesize
\begin{tabular}{lcc}
\toprule
\multirow{2}{*}{\textbf{Method}} & \textbf{VideoRefer-Bench\textsuperscript{Q}} (16) & \textbf{VideoInfer} (8) \\
\cmidrule(lr){2-2} \cmidrule(lr){3-3}
 & \textbf{Acc.} & \textbf{Acc. / Score} \\
\midrule
w/o STOM & 72.2 & 43.6 / 2.46\\
w/ \hspace{2pt} STOM & 74.0 & 46.7 / 2.60\\
\textcolor{gray}{w/ \hspace{2pt} oracle prop.} & \textcolor{gray}{75.1} & \textcolor{gray}{44.8 / 2.52} \\
\bottomrule
\end{tabular}
\vspace{-1mm}
\caption{Ablation on propagating visual prompts across the remaining frames. (16) represents 16 frames as input for inference.}
\label{ablation:prop}
\end{table}

\begin{table}[t]
\centering
{
\footnotesize
\begin{tabular}{lccccc}
\toprule
\multirow{3}{*}{\textbf{Method}} & \multicolumn{3}{c}{\textbf{ReVOS}~\cite{yan2024visa}} & \multicolumn{2}{c}{\textbf{ReasonSeg}\textsubscript{\textit{val}}~\cite{lai2024lisa}} \\ 
\cmidrule(lr){2-4} \cmidrule(lr){5-6}
 & $\mathcal{J\&F}$ & $\mathcal{J}$ & $\mathcal{F}$ & gIoU & cIoU \\ 
\midrule
LISA-7B~\cite{lai2024lisa} & 40.9 & 39.1 & 42.7 & 52.9 & 54.0 \\
VISA-7B~\cite{yan2024visa} & 46.9 & 44.9 & 49.0 & 52.7 & 57.8 \\ 
VideoLISA-3.8B~\cite{bai2025one} & 48.4 & 46.1 & 50.7 & 61.4 & 67.1 \\
\midrule
RGA3-7B-SAM & 55.4 & 53.4 & 57.4 & 68.5 & 68.6 \\
RGA3-7B-SAM2 & 58.0 & 55.9 & 60.0 & 68.7 & 70.2 \\ 
\bottomrule
\end{tabular}
}
\vspace{-1mm}
\caption{Comparison on the choice of the grounding module.}
\label{supp:sam}
\end{table}

\begin{table}[h]
\vspace{-6pt}
\centering
\footnotesize
\begin{tabular}{cccc}
\toprule
\textbf{Component} & Latency (s/sample) & FPS & Memory \\
\midrule
STOM & 0.483 & 43.0 & 3.8G \\
SAM2~(Hiera-L) & 4.543 & 35.6 & 5.4G\\
\bottomrule
\end{tabular}
\vspace{-1mm}
\caption{Latency and memory usage of STOM and SAM2.}
\label{tab:speed}
\end{table}
    
\section{Conclusion}
\label{conclusion}
In this work, we introduce RGA3, a multi-modal interactive VideoLLM that unifies video referring and grounding tasks in a multi-round conversational manner. 
By incorporating the Spatial-Temporal Overlay Module (STOM), our model effectively processes versatile visual prompts at any timestamp, enhancing user interaction and fine-grained object reasoning in videos. 
Furthermore, we introduce VideoInfer, a manually curated video instruction dataset designed to challenge Video LLMs with deep reasoning and inference-based question-answering, surpassing existing datasets in complexity and reasoning depth. 
Through extensive evaluations across VideoInfer as well as 11 existing benchmarks, we demonstrate the superior performance of RGA3 in both referring question-answering and reasoning-based segmentation tasks. 
We publicly release our dataset, code, and web demo to facilitate future advancements in video understanding. 
We hope that RGA3 and VideoInfer will serve as a strong foundation for future research, pushing the boundaries of object-centric video comprehension.

\clearpage
\setcounter{page}{1}
\maketitlesupplementary

\section{More Implementation Details}

\vspace{3pt} \noindent \textbf{Training Data Composition.}
The complete list of used datasets in training is presented in Tab.~\ref{supp:train_set}. 
For the datasets with overlapping, we select the disjoint samples during training, such as ViP-LLaVA-Instruct and LLaVA-150k.
\begin{table}[h]
\centering
\scalebox{0.9}{
\small
\begin{tabular}{lll}
\toprule
\textbf{Task} & \textbf{Datasets} & \textbf{\# Samples} \\
\midrule
\rowcolor{gray!10}
\multicolumn{3}{l}{\textit{Image Segmentation}} \\
\addlinespace[0.4em]
\multirow{4}{*}{Semantic Seg.} & ADE20k~\cite{zhou2017scene} & 20.2k \\
        & COCO-Stuff~\cite{caesar2018coco} & 118.3k \\
        & Pascal-Part~\cite{chen2014detect} & 4.3k \\
        & PACO-LVIS~\cite{ramanathan2023paco}  & 4.6k \\
\multirow{4}{*}{Referring Seg.} & RefCOCO+~\cite{kazemzadeh2014referitgame}  & 17k \\
          & RefCOCO~\cite{kazemzadeh2014referitgame}  & 22k \\
          & RefCOCOg~\cite{mao2016generation}  & 17k \\
          & RefCLEF~\cite{kazemzadeh2014referitgame}  & 18k \\
Reasoning Seg. & ReasonSeg~\cite{lai2024lisa}  & 0.2k \\
\midrule
\rowcolor{gray!10}
\multicolumn{3}{l}{\textit{Video Segmentation}} \\
\addlinespace[0.4em]
VOS & YoutubeVOS~\cite{seo2020urvos}  & 3.5k \\
         & Ref-Youtube-VOS~\cite{seo2020urvos}  & 3.5k \\
Referring VOS  & MeViS~\cite{ding2023mevis}  & 1.6k \\
         & Ref-DAVIS~\cite{khoreva2019video}  & 5.3k \\
Reasoning VOS & ReVOS~\cite{yan2024visa}  & 0.6k \\
\midrule
\rowcolor{gray!10}
\multicolumn{3}{l}{\textit{Image QA}} \\
\addlinespace[0.4em]
VQA & LLaVA-150k~\cite{liu2024visual}  & 150k \\
         & Osprey-Conv~\cite{yuan2024osprey}  & 30k \\
Referring VQA & Osprey-Desc~\cite{yuan2024osprey}  & 60k \\
         & ViP-LLaVA-Instruct~\cite{cai2024vip}  & 216k \\
\rowcolor{gray!10}
\midrule
\multicolumn{3}{l}{\textit{Video QA}} \\
\addlinespace[0.4em]
\multirow{6}{*}{VideoQA}        & LLaVA-Video-OE~\cite{zhang2024video}  & 960k \\
        & LLaVA-Video-MC~\cite{zhang2024video}  & 196k \\
        & NeXT-QA-OE~\cite{xiao2021next}  & 17k \\
        & NeXT-QA-MC~\cite{xiao2021next}  & 17k \\
        & ActivityNetQA~\cite{yu2019activitynet}  & 24k \\
        & PerceptionTest~\cite{patraucean2023perception}  & 2.4k \\
Referring VideoQA & VideoInfer~(Ours) & 20k \\
\bottomrule
\end{tabular}
}
\vspace{-2mm}
\caption{The detailed list of training datasets. For some datasets, we only use a subset of them, such as LLaVA-Video, Osprey, and ViP-LLaVA. The sampling rate is listed in the training script.}
\label{supp:train_set}
\end{table}

\vspace{3pt} \noindent \textbf{Training Details.}
We utilize the dynamic resolution for Qwen2.5-VL~\cite{Qwen2.5-VL} models, the max pixels of videos are set as 320$\times$28$\times$28 for 8 frames, and the max pixels of a single image are set as 1280$\times$28$\times$28. 
Images exceeding the above max pixels will be resized while maintaining their aspect ratio to fit.

\vspace{3pt} \noindent \textbf{Evaluation Protocols.} The results of MeViS \textit{val} and Ref-YouTube-VOS are evaluated through the online server. 
On VideoRefer-Bench\textsuperscript{Q}, our method processes 16 input frames. 
For VideoInfer, we utilize GPT-4o-2024-11-20 to evaluate Accuracy (Acc.) and Score, following the same prompt strategy as Video-ChatGPT~\cite{maaz2023video}. 
To evaluate region-feature based models, we convert the visual prompt~(RGBA image) into the mask according to the alpha channel and then input the mask with visual input and question into these models to generate the response.  

\section{More Experimental Results}

\subsection{Quantitative Results}
\begin{table}[h]
\centering
\setlength{\tabcolsep}{2.0pt}
\resizebox{0.48\textwidth}{!}{
\begin{tabular}{lccc}
\toprule
\multirow{1}{*}{\textbf{Method}}& \textbf{Perception Test}~\cite{patraucean2023perception} & \textbf{MVBench}~\cite{li2024mvbench} & \textbf{NExT-QA}~\cite{xiao2021next} \\
\midrule
\rowcolor{gray!10}
\multicolumn{4}{c}{\textit{Generalist Models}} \\
\addlinespace[0.2em]
LLaVA-OV-7B~\cite{li2024llava-one} & - & 56.7 & 79.4 \\ 
VideoLLaMA2.1-7B~\cite{cheng2024videollama} & 54.9 & 57.3 & 75.6 \\ 
LLaVA-Video-7B~\cite{zhang2024video} & 67.9 & 58.6 & 83.2 \\ \midrule
\rowcolor{gray!10}
\multicolumn{4}{c}{\textit{Specialist Models}} \\
\addlinespace[0.2em]
Artemis~\cite{qiu2024artemis} & 47.1 & 34.1 & - \\ 
VideoRefer-7B~\cite{yuan2024videorefer} & 56.3 & 59.6 & - \\ 
\arrayrulecolor{gray}\hdashline
\addlinespace[0.3em]
\textbf{RGA3-7B (Ours)} & 68.7 & 63.8 & 75.3 \\ \bottomrule
\end{tabular}
}
\vspace{-2mm}
\caption{Comparison on general video question-answering tasks.}
\label{supp:general_qa}
\end{table}

\vspace{3pt} \noindent \textbf{Results on General VideoQA Benchmarks.}
In addition to the referring video question-answering benchmarks, we also evaluate our architecture on general VideoQA QA tasks without visual prompts as inputs, through the LMMs-Eval toolkit~\cite{zhang2024lmms}. 
As shown in Tab.~\ref{supp:general_qa}, our model is comparable to popular general VideoQA models while possessing the ability to perform interactive referring and grounding in object-centric scenarios.

\begin{table}[th]
{\setlength{\tabcolsep}{4.0pt}
\centering
\resizebox{0.48\textwidth}{!}{
\begin{tabular}{l|cc|cc|cc|cc}
\toprule
& \multicolumn{2}{c|}{\textbf{val}} & \multicolumn{6}{c}{\textbf{test}} \\
\cmidrule(lr){2-3} \cmidrule(lr){4-9}
\textbf{Method} & \multicolumn{2}{c}{overall} & \multicolumn{2}{|c|}{short query} & \multicolumn{2}{c|}{long query} & \multicolumn{2}{c}{overall} \\
\cmidrule(lr){2-3} \cmidrule(lr){4-5} \cmidrule(lr){6-7} \cmidrule(lr){8-9}
& gloU & cloU & gloU & cloU & gloU & cloU & gloU & cloU \\
\midrule
\multicolumn{1}{l|}{LISA-7B~\cite{lai2024lisa}} & 52.9 & 54.0 & 40.6 & 40.6 & 49.4 & 51.0 & 47.3 & 48.4 \\
\multicolumn{1}{l|}{LISA-13B~\cite{lai2024lisa}} & 56.2 & 62.9 & 44.3 & 42.0 & 54.0 & 54.3 & 51.7 & 51.1 \\
\multicolumn{1}{l|}{VISA-7B~\cite{yan2024visa}} & 52.7 & 57.8 & - & - & - & - & - & - \\
\multicolumn{1}{l|}{VideoLISA-3.8B~\cite{bai2025one}} & 61.4 & 67.1 & 43.8 & 42.7 & 56.9 & 57.7 & 53.8 & 54.4 \\
\multicolumn{1}{l|}{LISA++-7B~\cite{yang2023lisa++}} & 64.2 & 68.1 & 49.6 & 51.1 & 59.3 & 61.7 & 57.0 & 59.5 \\
\midrule
 \rowcolor{LightSkyBlue!15}
\multicolumn{1}{l|}{RAG3-3B~(Ours)} & 65.4 &	68.5 &	58.5 & 54.2	& 62.3 & 65.8 & 61.4 & 63.3  \\
 \rowcolor{LightSkyBlue!15}
\multicolumn{1}{l|}{RAG3-7B~(Ours)} &  \textbf{68.7}  &  \textbf{70.2}  &  \textbf{58.7}  & \textbf{54.1}  & \textbf{68.5}  & \textbf{72.1}  &  \textbf{66.1} & \textbf{68.3}  \\
\bottomrule
\end{tabular}
}
\vspace{-2mm}
\caption{Comparision on validation and test set of ReasonSeg~\cite{lai2024lisa} for image-level reasoning object segmentation.}
\label{tab:image_reason_seg}
}
\end{table}
\begin{table}[t]
\setlength{\tabcolsep}{2.0pt}
\centering
\resizebox{0.48\textwidth}{!}{
\begin{tabular}{l|ccc|ccc|cc}
\toprule
\multirow{2}{*}{\textbf{Method}} & \multicolumn{3}{c|}{\textbf{refCOCO}~\cite{kazemzadeh2014referitgame}} & \multicolumn{3}{c|}{\textbf{refCOCO+}~\cite{kazemzadeh2014referitgame}} & \multicolumn{2}{c}{\textbf{refCOCOg}~\cite{mao2016generation}} \\ 
\cmidrule(lr){2-4} \cmidrule(lr){5-7} \cmidrule(lr){8-9} 
 & val & testA & testB & val & testA & testB & val(U) & test(U) \\ 
 \midrule
LISA-7B~\cite{lai2024lisa} & 74.1 & 76.5 & 71.1 & 62.4 & 67.4 & 56.5 & 66.4 & 68.5 \\
VISA-7B~\cite{yan2024visa} & 72.4 & 75.5 & 68.1 & 59.8 & 64.8 & 53.1 & 65.5 & 66.4 \\
VideoLISA-3.8B~\cite{bai2025one} & 73.8 & 76.6 & 68.8 & 63.4 & 68.8 & 56.2 & 68.3 & 68.8 \\ 
\midrule
 \rowcolor{LightSkyBlue!15}
RGA3-3B~(Ours) & 78.9	& 81.1	& 75.0 & 71.3 &	77.1 &	63.5 & 74.7	& 75.0  \\
 \rowcolor{LightSkyBlue!15}
RGA3-7B~(Ours) & \textbf{79.7} & \textbf{82.6} &	\textbf{76.0} & 	\textbf{73.5} &	\textbf{78.6} &	\textbf{67.0} & \textbf{76.2} &	\textbf{75.9} \\ 
\bottomrule
\end{tabular}
}
\vspace{-2mm}
\caption{Comparison on image-level referring object segmentation datasets.}
\label{tab:image_ref}
\end{table}

\vspace{3pt} \noindent \textbf{Results on Image Segmentation Benchmarks}
For image segmentation evaluation, we utilize gIoU (average per-image IoUs) and cIoU (cumulative intersection over union) on reasoning-based benchmark ReasonSeg~\cite{lai2024lisa} and cIoU for referring-based benchmark refCOCO(+/g)~\cite{kazemzadeh2014referitgame,mao2016generation}. 
As shown in Tab.~\ref{tab:image_reason_seg}, RGA3-7B outperforms the state-of-the-art method LISA++-7B~\cite{yang2023lisa++} on the ReasonSeg benchmark by a large margin.
Moreover, on general image semantic segmentation benchmarks, such as refCOCO, RGA3-7B still outperforms recent MLLM-based methods, which indicates the strong general grounding ability of RGA3.

\vspace{3pt} \noindent \textbf{Robustness in Extremely Long Videos.}
Our work primarily addresses object-centric video tasks, which typically involve short video durations ({\em e.g.,} VideoRefer-Bench\textsuperscript{Q} comprises a few-second clips sourced from DAVIS or MeVIS). 
Although our VideoInfer incorporates longer clips (sub-minute duration) from LVOS and TAO, we acknowledge the necessity for robustness in ultra-long video.
Due to the lack of appropriate benchmarks, we evaluated RGA3 on the validation set of the LongVideoBench with different duration groups: 
\begin{table}[h]
\vspace{-6pt}
\centering
\resizebox{0.48\textwidth}{!}{
\begin{tabular}{lcccc}
\toprule
Duration & $(8s, 15s]$ & $(15s, 60s]$ & $(180s, 600s]$ & $(900s, 3600s]$ \\
\midrule
Accuracy & 72.5 & 70.9 & 57.3 & 46.3 \\
\bottomrule
\end{tabular}
}
\vspace{-2mm}
\caption{Performance on the validation set of LongVideoBench.}
\label{tab:longvideobench_val}
\end{table}

\vspace{3pt} \noindent \textbf{More Ablations.}
The improvement over the previous state-of-the-art methods on video referring segmentation and question-answering is mainly from the base MLLM, the proposed STOM module, and the dataset composition in training. 
Additionally, we find that under the current training strategy, the performance of the individual task will decrease compared to training separately in most cases. 
We think this should be further addressed through multi-stage training or more diverse prompting. 

\begin{table}[h]
\vspace{-6pt}
\centering
{
\setlength{\tabcolsep}{3.0pt}
\resizebox{0.48\textwidth}{!}{
\begin{tabular}{@{}ccccc@{}cc@{}}
\toprule
\multirow{2}{*}{\textbf{Model}} & \multirow{2}{*}{\textbf{Training Data}} & \multirow{2}{*}{\textbf{MLLM}} &\multirow{2}{*}{\textbf{Size}} & \multirow{2}{*}{\textbf{Modules}} & \textbf{VideoRefer} & \textbf{ReasonVOS} \\
\cmidrule(lr){6-6} \cmidrule(lr){7-7} 
 & & & & & Acc. & $\mathcal{J} \& \mathcal{F}$  \\
\midrule
VideoRefer & QA & SigLIP-Qwen2 & 7B& - & 71.9 & - \\
VideoLISA & Seg & Phi-3-V &3.8B& SAM & - & 45.1 \\
\arrayrulecolor{gray}\hdashline
\multirow{5}{*}{Ours} & Seg & \multirow{5}{*}{QwenVL-2.5} &3B &SAM  & - & 47.9  \\
 & Seg & &3B & \multirow{1}{*}{SAM\rlap{2}} & - & 48.7  \\
 &  Seg+QA & & 3B &SAM\rlap{2} & 62.3 &  51.7 \\
 &Seg+QA & & 3B & SAM2 +STOM & 66.6 & 51.7  \\
  &Seg+QA & & 7B & SAM2 +STOM & 74.0 & 53.6  \\
\bottomrule
\end{tabular}
}
}
\vspace{-2mm}
\caption{Additional ablations on the design choices.}
\label{rebuttal:ablation}
\end{table}

\subsection{Failure Case and Future Work}
\begin{figure}[!t]
\begin{center}
\includegraphics[width=\linewidth]{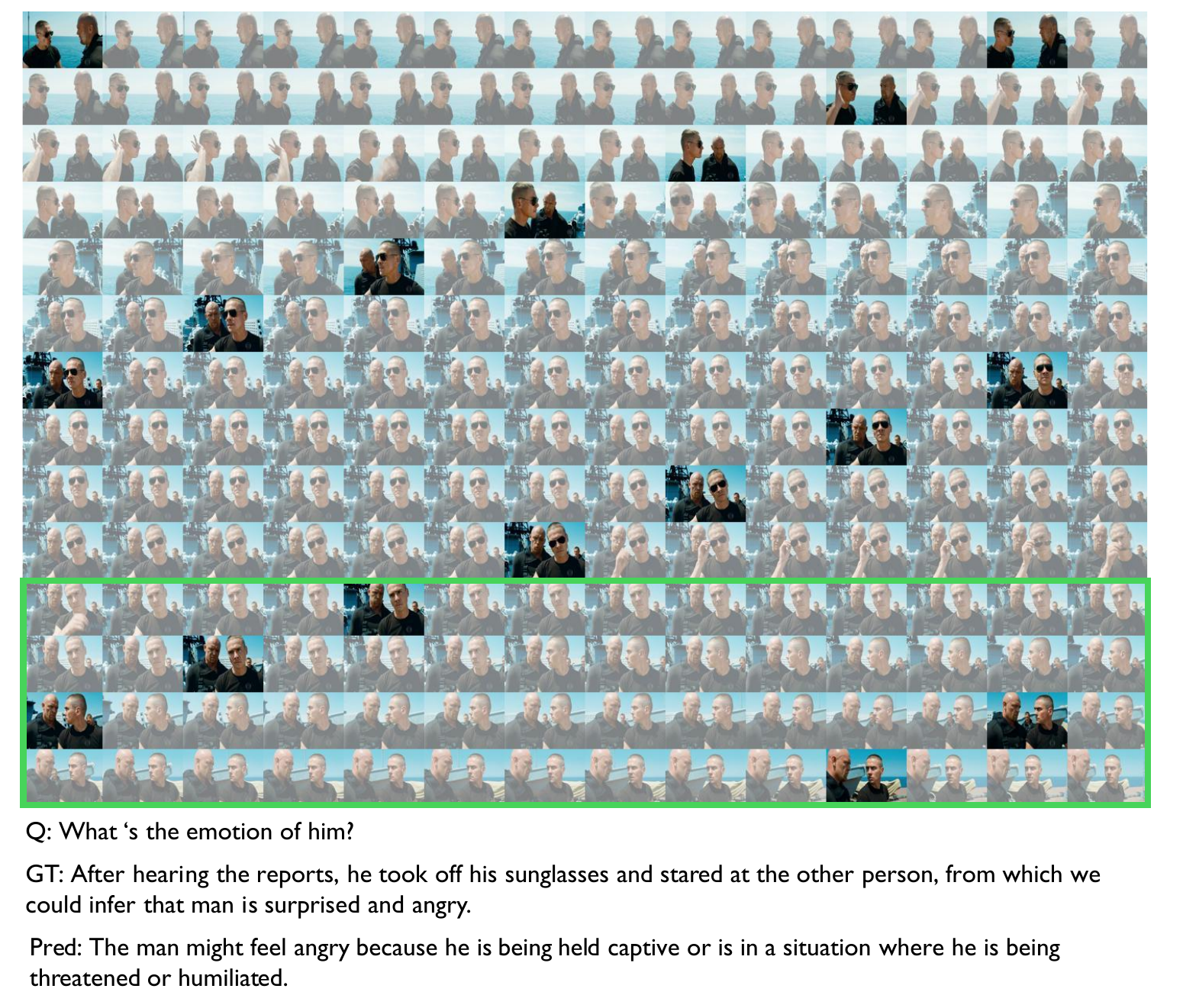}
\end{center}
\vspace{-5mm}
\caption{Failure case on VideoInfer dataset. The frames with grey masks are not selected as input to RGA3. The green box frames the video content which the man on the left reports something to the man on the right. `GT' is the ground truth, and `Pred' is the prediction of RGA3.}
\label{fig:failure}
\end{figure}

In practice, due to computational limitations, we restrict RGA3’s input to 16 frames per video~(Other existing object-centric VideoLLMs also suffer from this limitation).
However, in very long videos, this frame selection introduces large temporal gaps, potentially omitting critical contextual information.
Our VideoInfer dataset introduces videos that contain over 1,000 frames, yet the existing models can not process the whole sequence due to computational limitations. 
For instance, as shown in Fig.~\ref{fig:failure}, the raw video contains over 1,000 frames, yet only 16 frames are used as input. 
With such sparse frame sampling, the model struggles to capture a coherent sequence of events.

\begin{figure*}[tbph]
    \centering
    \begin{subfigure}[b]{0.48\textwidth}
        \centering
        \includegraphics[width=\textwidth]{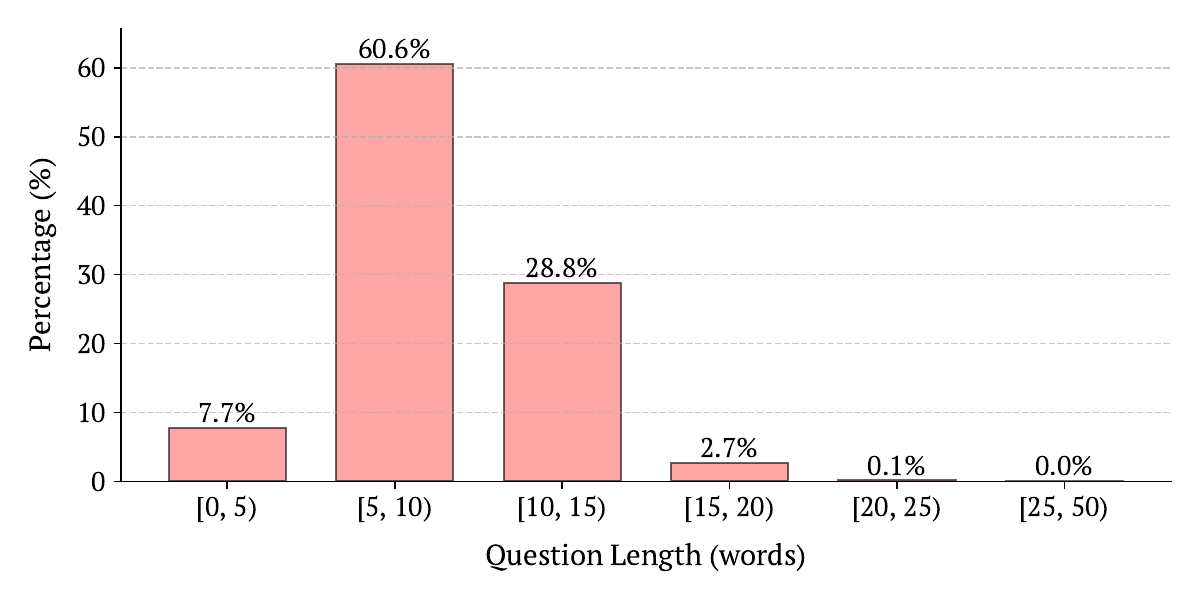}
        \caption{Histogram of question word counts.}
    \end{subfigure}
    \hfill
    \begin{subfigure}[b]{0.48\textwidth}
        \centering
        \includegraphics[width=\textwidth]{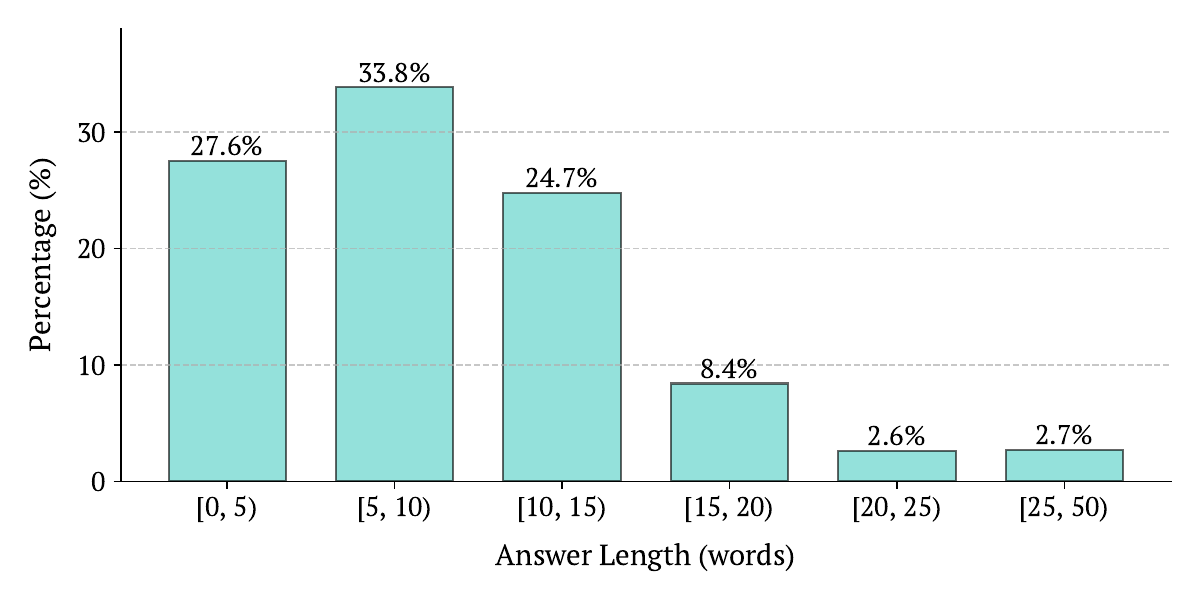}
        \caption{Histogram of answer word counts.}
    \end{subfigure}
    
    \vskip 20pt
    
    \begin{subfigure}[b]{0.48\textwidth}
        \centering
        \includegraphics[width=\textwidth]{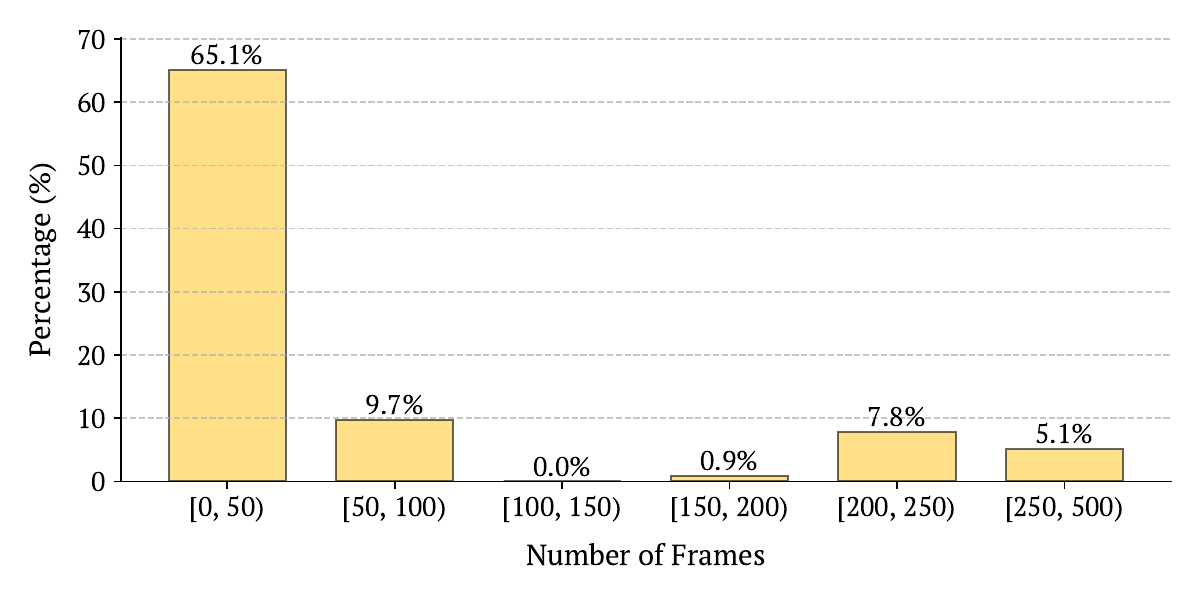}
        \caption{Histogram of frames per video counts.}
    \end{subfigure}
    \hfill
    \begin{subfigure}[b]{0.48\textwidth}
        \centering
        \includegraphics[width=\textwidth]{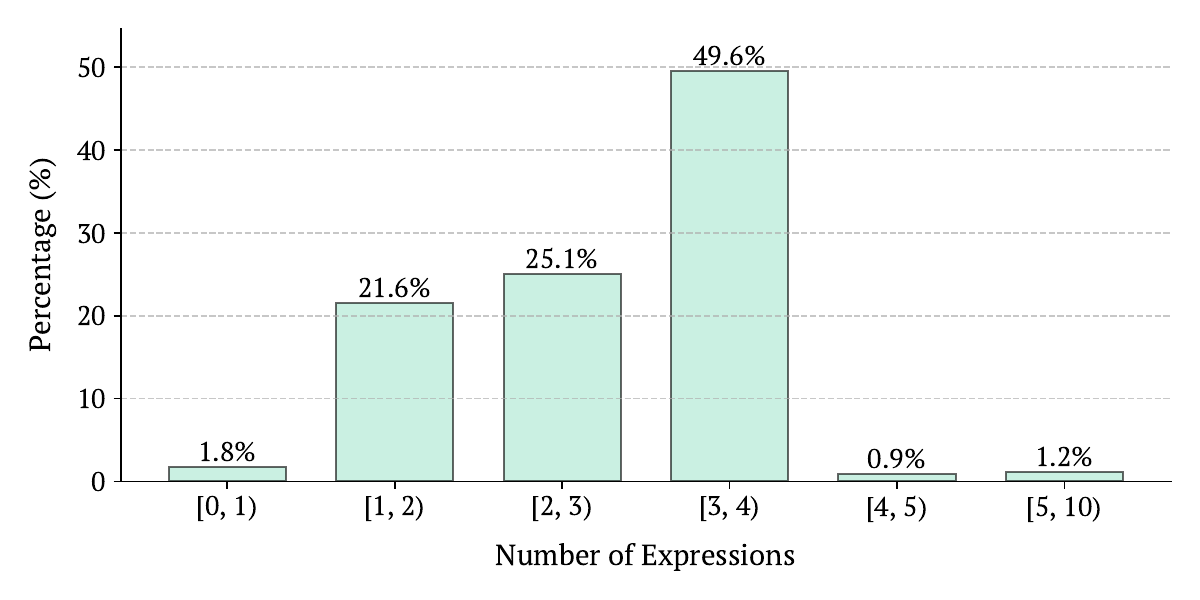}
        \caption{Histogram of objects per video counts.}
    \end{subfigure}
    
    \caption{Visualization of statistics of the test split of VideoInfer.}
    \label{fig:statistics}
\end{figure*}

\noindent In this specific case, the moment when the left person reports to the right person is skipped, leading to an incorrect prediction.
This issue cannot be naively addressed by extracting just one or a few visual tokens per frame, as object-level information must be preserved across frames to enable accurate object-centric reasoning. 
Therefore, handling long-form object-centric video reasoning remains a challenging open problem, particularly in transforming spatial and temporal detailed object-centric information into a reasonable number of tokens. 
We plan to explore solutions further to enhance object-centric reasoning in long videos in our future work.

\section{Discussion and Visualizations}


\subsection{Potential Information Loss}

The STOM module blends prompts onto original frames with \textbf{transparency} through alpha blending, so that the objects will not be completely occluded, and the features can be reserved.

\subsection{Visualization of VideoInfer Benchmark}
As shown in Fig.~\ref{fig:statistics}, we visualize the statistics of the test split in VideoInfer. The questions range from 3 to 50 words in length, with an average of 8.4 words. Answers vary between 1 and 75 words, averaging 8.7 words. The number of frames per sample spans from 7 to over 2000, with a mean of 189.5 frames. For objects, the count ranges from 1 to 8, averaging 2.3 objects of interest per video.

\end{document}